\documentclass[lettersize, journal]{IEEEtran}
\ifCLASSINFOpdf
    \usepackage[pdftex]{graphicx}
    \usepackage{amsmath,amsfonts}
    \usepackage{array}
    \usepackage[caption=false,font=normalsize,labelfont=sf,textfont=sf]{subfig}
    \usepackage{textcomp}
    \usepackage{stfloats}
    \usepackage{url}
    \usepackage{verbatim}
    \usepackage{graphicx}
    \usepackage{cite}
    \usepackage{booktabs} 
    \usepackage{balance}
    \usepackage[breaklinks=true,colorlinks, citecolor=blue]{hyperref}
    \usepackage{multirow} 
    \usepackage{bbding}
    \usepackage{pifont}
    \usepackage{wasysym}
    \usepackage{amssymb}
    \usepackage{hyperref}
    \usepackage{bbm}
    \usepackage{threeparttable} 
    \usepackage{booktabs}
    \usepackage{xcolor}
    \usepackage{blindtext}
    \usepackage{bm}
    \usepackage{amssymb,mathtools}
    \usepackage[linesnumbered, ruled]{algorithm2e}
    \usepackage[table]{xcolor}
\else
\fi
\begin{document}
%
\title{DeH4R: A Decoupled and Hybrid Method for Road 

Network Graph Extraction}
%
%
%

\author{Dengxian~Gong,
        Shunping~Ji,~\IEEEmembership{Senior Member,~IEEE}
        
\thanks{
This work was supported by the National Natural Science Foundation of China under Grant 42571412. (Corresponding author: Shunping Ji)

Dengxian Gong and Shunping Ji are with the School of Remote Sensing and
Information Engineering, Wuhan University, Wuhan 430079, China (e-mail:
gooodx@whu.edu.cn; jishunping@whu.edu.cn).}
}

\maketitle

\begin{abstract}
The automated extraction of complete and precise road network graphs from 
remote sensing imagery remains a critical challenge in geospatial computer vision. 
Segmentation-based approaches, 
while effective in pixel-level recognition, struggle to maintain topology fidelity after vectorization post-processing. 
Graph-growing methods build more topologically faithful graphs but suffer from computationally prohibitive iterative ROI cropping. 
Graph-generating methods first predict global static candidate road network vertices, 
and then infer possible edges between vertices. 
They achieve fast topology-aware inference, but limits the dynamic insertion of vertices. 
To address these challenges, 
we propose DeH4R, a novel hybrid model that combines graph-generating efficiency and graph-growing dynamics. 
This is achieved by decoupling the task into 
candidate vertex detection, 
adjacent vertex prediction, 
initial graph construction, 
and graph expansion. 
This architectural innovation enables dynamic vertex (edge) insertions 
while retaining fast inference speed and enhancing both topology fidelity and spatial consistency. 
Moreover, it is simple and straightforward to implement.
Comprehensive evaluations on the CityScale and SpaceNet benchmarks demonstrate state-of-the-art (SOTA) performance.
DeH4R outperforms the prior SOTA graph-growing method RNGDet++ by \textbf{4.66} APLS and \textbf{10.18} IoU on CityScale, 
while being approximately 10$\times$ faster. The code is publicly available at \url{https://github.com/7777777FAN/DeH4R}.
\end{abstract}


%
\IEEEpeerreviewmaketitle

\section{Introduction}
\label{sec:intro}
%
%
%
%
\IEEEPARstart{R}{oad} network graphs constitute fundamental vector data
for applications such as autonomous driving, navigation,
and urban planning. Current road network graph products
primarily rely on time-consuming and costly manual annotation. 
Automating the extraction of road network graphs from
high-resolution satellite and aerial imagery has emerged as
a critical research focus to enhance production efficiency in
geospatial intelligence applications.

\begin{figure}
  \centering
  \includegraphics[width=\linewidth]{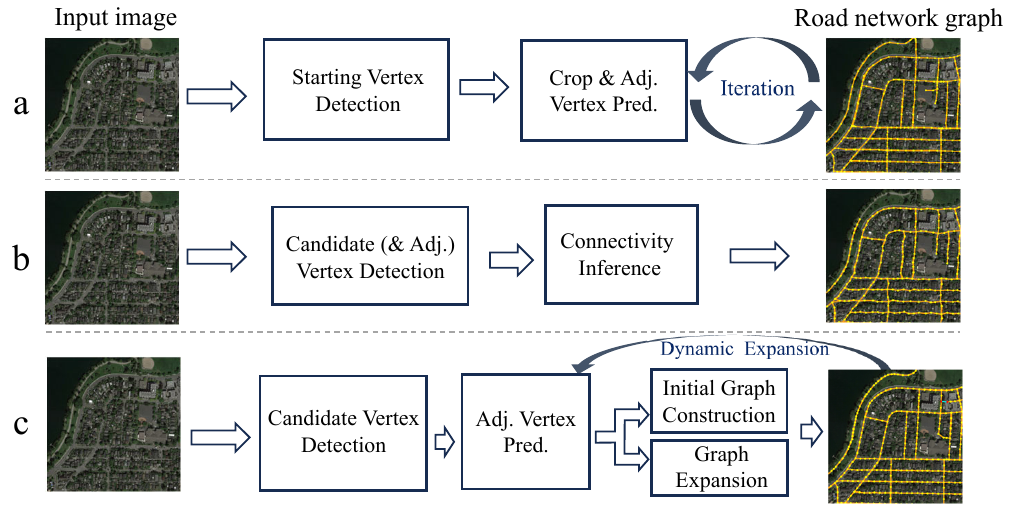}
   \caption{Pipelines of existing methods and our method. 
   (a) Graph-growing methods are computationally expensive 
   due to its iterative ROI cropping and full-model forward passes.
   (b) Graph-generating methods show great potential in parallel inference,
   but existing methods infer edges between a fixed set of candidate vertices 
   (or candidate vertices and their adjacent vertices extracted in one-step),  
   precluding dynamics for vertex expansion.
   (c) Our approach decouples the task into candidate vertex detection, 
   adjacent vertex prediction, initial graph construction and graph expansion,
   thus enabling dynamics and high inference efficiency.
   }
   \label{fig:teaser}
\end{figure}

Contemporary extraction approaches primarily fall into three categories: 
segmentation-based methods, 
graph-growing methods,
and graph-generating methods.
Segmentation-based techniques \cite{mnih2010learning, zhang2018road, zhou2018d, mattyus2017deeproadmapper, mei2021coanet, batra2019improved} typically generate pixel-level masks followed by 
post-processing pipelines (e.g., morphological operation, path-finding) to derive road network graphs. 
However, these masks frequently exhibit geometric imperfections such as fractures and distortions, 
which compromise the topology fidelity of the final road network graph. 

Graph-based methods directly extract road network graphs from imagery, ensuring superior topological fidelity. 
These approaches are further subdivided into iterative graph-growing algorithms and non-iterative graph-generating algorithms.
Graph-growing algorithms \cite{bastani2018roadtracer, tan2020vecroad, li2019topological, xu2023rngdet++, sotiris2023mastering} frame the problem as a sequential next-point prediction task, 
requiring full model forward inference over highly overlapping regions of interest (ROIs) at each prediction. 
While achieving exceptional topological accuracy, 
their computational overhead remains prohibitive. 
In contrast, graph-generating algorithms \cite{he2020sat2graph, bahl2022single, xu2022csboundary, yang2023topdig, hetang2024segment} 
isolate vertex detection and edge inference, 
enabling parallel computation for accelerated processing. 
Nevertheless, their reliance on the global static candidate vertex set from the vertex detection stage 
inherently limits dynamic vertex replenishment and topological flexibility.

This analysis motivates our core inquiry: 
Can we unify graph-growing dynamics
with graph-generating efficiency, 
thereby advancing road network graph extraction performance?

In response, we propose DeH4R, a novel hybrid model that employs a four-stage decoupled workflow.
Specifically, DeH4R leverages Segment Anything Model (SAM) \cite{kirillov2023segment, ravi2024sam} for accurate candidate vertex localization,
adopts a DETR-like architecture \cite{carion2020end} to predict adjacent vertices for all candidates,
decodes extracted vertices and their predicted adjacent vertices into an initial graph using the graph-generating idea,
and dynamically expand the graph into a complete network borrowing the idea of graph-growing.
With this decoupled strategy, 
DeH4R is capable of building a road network graph that spatially aligns well with the ground truth 
via graph-generating and graph-growing while being approximately 10$\times$ faster than
the prior state-of-the-art (SOTA) graph-growing method.

Our key contributions are:
\begin{itemize}
  \item We propose the DeH4R model which unifies graph-growing dynamics with graph-generating efficiency through a decoupling strategy, effectively harnessing their complementary strengths.
  \item The proposed DeH4R model achieves new SOTA results with a significant improvement over previous methods and exceptional inference speed on two mainstream public benchmarks.
\end{itemize}

\section{Related Work}
\label{sec:related}
\subsection{Segmentation-Based Methods}

Segmentation-based approaches typically commence with semantic segmentation to obtain pixel masks, 
followed by vectorization post-processing pipelines (e.g.,, morphological thinning \cite{zhang1984fast}, path search \cite{hart1968formal}, 
and graph simplification \cite{douglas1973algorithms}) to derive road network graphs. 
The pioneering work \cite{mnih2010learning} applied deep learning to road surface mask extraction.
Subsequently, \cite{zhang2018road} enhanced mask prediction performance by introducing skip connections \cite{he2016deep}, 
while D-LinkNet \cite{zhou2018d} further combined skip connections with dilated convolutions to improve segmentation accuracy.
Later advancements has focused on topology refinement: DeepRoadMapper \cite{mattyus2017deeproadmapper} formalized edge inference as a shortest-path optimization framework, 
leveraging the A* algorithm \cite{hart1968formal} to recover missing topological connections in initial road network graphs. 
Improved-Road \cite{batra2019improved} incorporated road directional priors during graph construction, 
whereas CoANet \cite{mei2021coanet} explicitly accounted for the ribbon-like geometry of roads through introducing strip convolution operators to enhance feature representation. 
$C^{2}Net$\cite{yang2024c2net} and CRNet \cite{11397410} improve road topology and connectivity in occluded regions by incorporating global-local context perception modules and semantic-spatial feature refinement mechanisms to suppress background interference.

\subsection{Graph-Based Methods}
Current graph-based approaches directly generate road network topologies through two main approaches: 
iterative graph-growing algorithms \cite{bastani2018roadtracer, tan2020vecroad, li2019topological, xu2023rngdet++, sotiris2023mastering}
and parallelizable graph-generating algorithms \cite{he2020sat2graph, bahl2022single, xu2022csboundary, yang2023topdig, hetang2024segment}. 
In both categories, semantic segmentation primarily serves to identify starting vertices or generate massive candidate vertices.

\textbf{Graph-Growing Algorithms.} 
RoadTracer \cite{bastani2018roadtracer} pioneered an iterative approach that 
trains a CNN-based decision network to directly generate road graphs from imagery. 
This network incrementally predicts movement decisions and directions within predefined regions of interest (ROIs), 
propagating paths with fixed step sizes. 
To address intersection drift issues in RoadTracer, 
VecRoad \cite{tan2020vecroad} introduced flexible step size and 
leveraged road keypoints alongside road surface segmentation masks to constrain coordinate regression, 
thereby enhancing connectivity and spatial precision. Recently, inspired by DETR \cite{carion2020end}, 
RNGDet++ \cite{xu2023rngdet++} reformulated adjacent vertex prediction as a set prediction task, 
employing Transformer architectures \cite{vaswani2017attention} to directly regress coordinates of multiple adjacent vertices. 
While our adjacent vertex prediction module draws inspiration from this set prediction strategy, 
we eschew its inefficient iterative inference mechanism.

\textbf{Graph-Generating Algorithms.} 
Sat2Graph \cite{he2020sat2graph} encodes the entire graph structure into a 19-dimensional tensor
which contains the coordinates of all candidate vertices and their possible adjacent vertices, 
enabling an elegant framework. 
Subsequent works \cite{bahl2022single, xu2022csboundary, he2022td, yang2023topdig, hetang2024segment}  
decomposed graph extraction into candidate vertex detection and edge inference stages. 
SAM-Road \cite{hetang2024segment} integrated the 
Segment Anything Model (SAM) \cite{kirillov2023segment} with parallel processing strategies, 
achieving state-of-the-art inference efficiency without compromising performance.
While our approach is inspired by SAM-Road \cite{hetang2024segment} and Sat2Graph \cite{he2020sat2graph}, 
we deliberately decouple the road network graph extraction workflow rather than adopting a mixed vertex detection
and adjacent vertex prediction architecture as in Sat2Graph
or direct edge prediction as in SAM-Road. 
This decoupled design is used to better introduce dynamic vertex insertion and improve the expansion ability of the graph.

\subsection{SAM and DETR-like Works}
DETR (Detection Transformer) \cite{carion2020end} and Segment Anything Model (SAM) \cite{he2020sat2graph, ravi2024sam}  have 
demonstrated transformative capabilities in object detection and image segmentation, respectively. 
In this study, 
we harness the pre-trained SAM2 \cite{ravi2024sam} encoder for accurate candidate vertex detection and 
adapt DETR's core design principles for adjacent vertex prediction.

DETR \cite{carion2020end} introduced 
a pioneering end-to-end framework for object detection, simplifying traditional pipelines. 
Its versatility has inspired various applications, 
such as LETR \cite{xu2021line} for line-based representation learning, 
\cite{can2021structured, liao2022maptr, liu2023vectormapnet, liao2024maptrv2, zhou2024himap} for high-definition map generation, 
and RNGDet \cite{xu2022rngdet,xu2023rngdet++} which extends DETR to road network graph extraction with direct adjacent vertex prediction. 

With exceptional zero-shot generalization capability, 
SAM \cite{kirillov2023segment} has been widely applied in domains such as medical image segmentation \cite{huang2024segment, zhang2024segment, ma2024segment}, 
remote sensing \cite{wang2023samrs, chen2024rsprompter}, and image inpainting \cite{yu2023inpaint}. 
In the context of road network graph extraction, 
the success of SAM-Road \cite{hetang2024segment} highlights SAM's great potential. 
Recently, the newer SAM2 \cite{ravi2024sam} has expanded SAM into the video segmentation domain and 
incorporated a multi-scale architecture which helps detect small object \cite{lin2017feature} like candidate vertices in road. 
Inspired by these advancements, 
we employ the encoder of SAM2 as the visual encoder in this study, 
leveraging its multi-scale features to facilitate our task.

\begin{figure*}
  \centering
    \includegraphics[width=\linewidth]{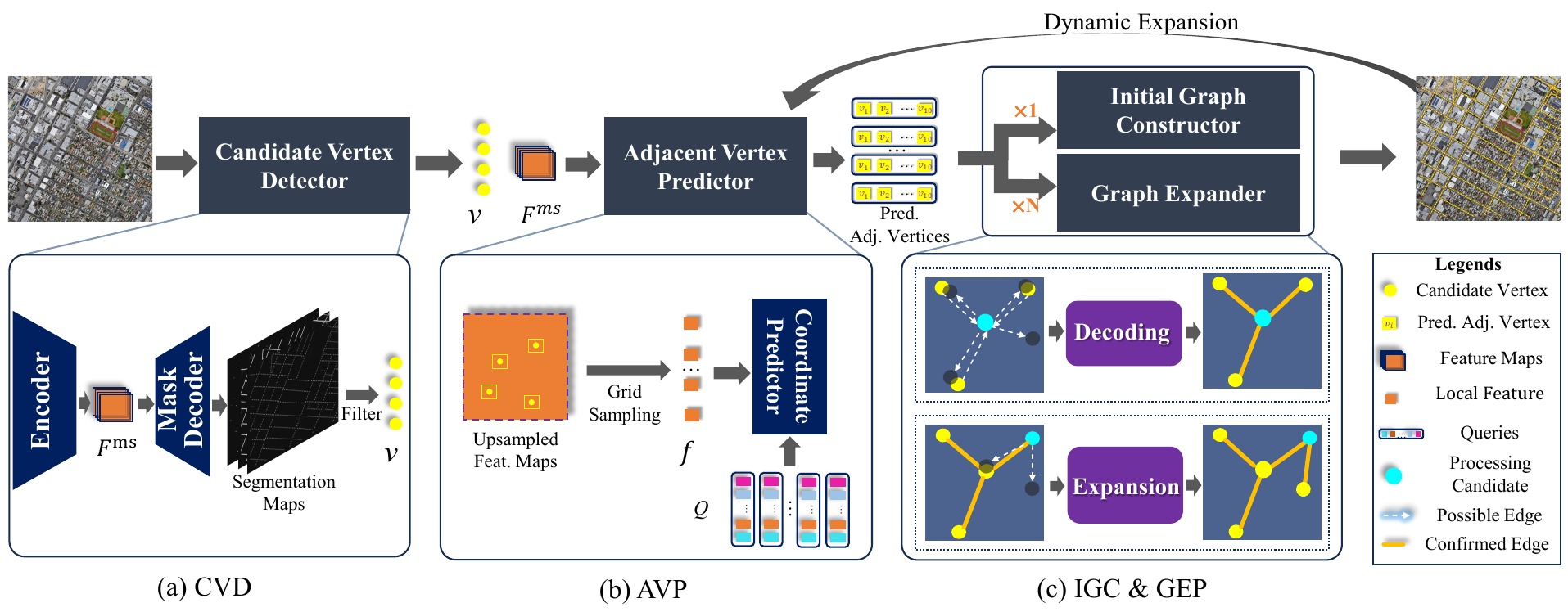}
    \caption{
    The pipeline of DeH4R.
		The upper row depicts the complete workflow of DeH4R, 
		aligned with our decoupling strategy: 
		Given an input image,
		CVD extracts features and detects candidate vertices, 
		AVP predicts adjacent vertices for all candidates, 
		IGC converts vertex predictions into a graph (inference) or generates labels for predicted vertices on-the-fly (training),
		and GEP expands the graph $N$ times to obtain a complete one.
		The lower row provides detailed explanations of CVD, AVP, IGC and GEP. 
    }
    \label{fig:archi}
  \hfill
\end{figure*}

\section{Method}
\label{sec:method}

The overall pipeline of DeH4R is illustrated in Fig. \ref{fig:archi}.
It converts roads in an input RGB image $I^{H \times W \times 3}$ into vertices $V = \{v_i\}$ and edges $E = \{e_j\}$ 
in a road network graph $G = \{V, E\}$, 
where each vertex $v_i$ represents a point in a road centerline, 
and each edge $e_j$ represents a segment of a road centerline. 
We decouple the graph extraction process into
candidate vertex detection, adjacent vertex prediction, initial graph construction and graph expansion,
thus we formulate our model as four main components: 
Candidate Vertex Detector (CVD) in Sec. \ref{sec:detector}, 
Adjacent Vertex Predictor (AVP) in Sec. \ref{sec:predictor}, 
Initial Graph Constructor (IGC) in Sec. \ref{sec:constructor}, 
and Graph Expander (GEP) in Sec. \ref{sec:expander}.

\subsection{Candidate Vertex Detector}
\label{sec:detector}
As depicted in Fig. \ref{fig:archi}a, the CVD takes a RGB image patch $I^{h \times w \times 3}$ as input and outputs the segmentation mask ${M}^{h \times w \times 3}$
which contains three segmentation maps: keypoint map, sampling point map, and road surface map.
Keypoints are road intersections and terminals, 
while sampling points are points sampled at regular intervals along road segments between 
any two keypoints (including the keypoints themselves).

The CVD consists of a Vision-Transformer-based (ViT) \cite{dosovitskiy2020image} visual encoder 
derived from SAM2 \cite{ravi2024sam} and a segmentation head.
The visual encoder consumes the input image patch $I^{h \times w \times 3}$ 
and learns hierarchical features $F$ = $\{f_i \mid 
i = 1, 2, 3, 4\}$ at 1/4, 1/8, 1/16, 1/32 spatial resolutions.
A feature pyramid network (FPN) \cite{lin2017feature} is employed 
to generate fused multi-scale image features $F^{ms} = \{f^{ms}_i \mid 
i = 1, 2, 3, 4\}$ with a unified channel dimension $d$, where only the $1/32$ scale lateral features are upsampled and fused with the $1/16$ scale lateral features through element-wise addition, whereas the $1/8$ and $1/4$ scales retain their original lateral representations.
The multi-scale features are cached for candidate vertices detection and adjacent vertices prediction.
The segmentation head is comprised of 4 transposed convolution layers 
and progressively upsamples the fused features to generate the segmentation mask ${M}^{h \times w \times 3}$.

We directly predict sampling points and use a morphology-based local minimum localization method (as specified in Alg. \ref{alg:localmin_math}) adopted by \cite{he2020sat2graph} 
to detect the majority of candidate vertices from keypoint and sampling point maps 
and then use NMS \cite{hetang2024segment} to reduce clustering of candidate vertices.
These vertices, generated by applying an NMS to road surface segmentation maps, serve only as a supplement. 
A final NMS removes any redundant vertices derived from the three segmentation maps
and outputs a set of candidate vertices $C=\{v_c\}$, $v\in \mathbb{R}^2$ where each vertex $v_c = {(x_i, y_i)}$.

\begin{figure}
  \centering
  \includegraphics[width=\linewidth]{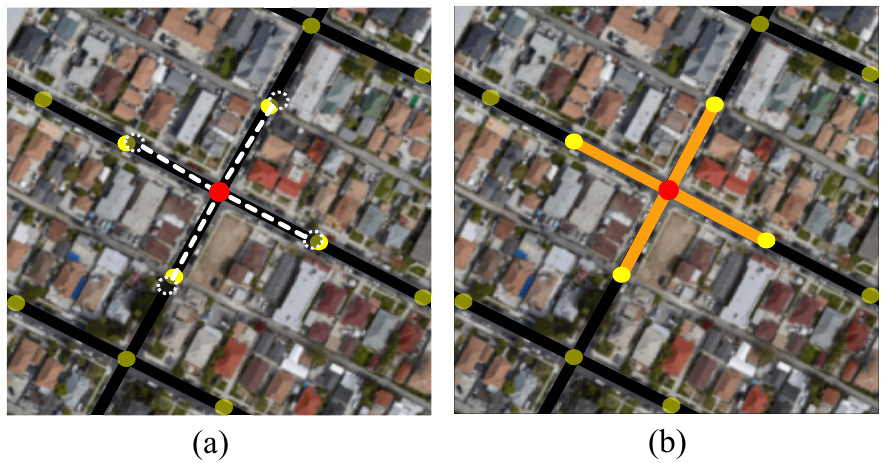}
  \caption{Schematic diagram of the effect before and after decoding. 
  (a) A discrepancy between candidate vertices and predicted adjacent vertices before decoding;
  (b) The connected candidate vertices after decoding.
  Yellow points are candidate vertices.
  Red point is the currently processing vertex.
  Dashed and white points and lines are predicted adjacent vertices and possible edges, respectively.
  Orange lines are confirmed edges.
  }
  \label{fig:decoding}
\end{figure}

\begin{figure}[t]
  \centering
  \includegraphics[width=\linewidth]{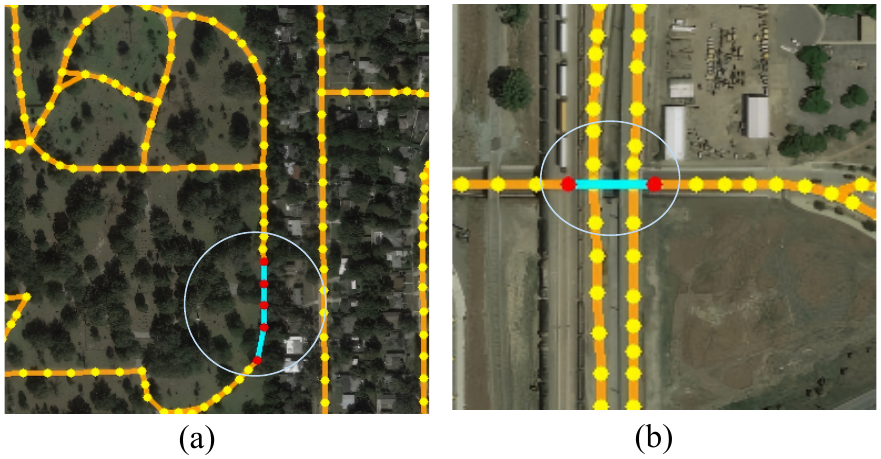}
   \caption{Examples of graph expansion. 
   (a) An insertion case where the predicted vertex is not in existing graph;
   (b) A merging case where the predicted vertex is close to an existing vertex and then merged.}
   \label{fig:expansion}
\end{figure}

\subsection{Adjacent Vertex Predictor}
	\label{sec:predictor}
	The design of AVP is inspired by RNGDet \cite{xu2022rngdet, xu2023rngdet++}.
	As illustrated in Fig. \ref{fig:archi}b, 
	after obtaining the candidate vertices $C = \{v_c\}$, 
	AVP takes as input the visual features $F^{ms}$ and $C$
	and outputs $N$ possible adjacent vertices for each candidate vertex. 

	AVP contains a coordinate regression head instantiated as 3 standard Transformer decoder layers and 1 linear layer.
	For each candidate vertex $v_c \in C$,
	a square ROI of size $l \times l$ is defined based on its coordinates on current patch,
	and we use grid sampling \cite{hetang2024segment,yang2023topdig} to 
	interpolate the multi-scale features $F^{ms}$ at each scale
	and then add them up point-wisely to form the final ROI feature $f^{roi}_c \in \mathbb{R}^{l \times l \times d}$, 
	where $d$ is the number of channels. 

	The local ROI feature $f^{roi}_c$, along with $N_q$ ($N_q = N$) learnable queries $q$ ($q \in \mathbb{R}^d$), 
	are fed into the coordinate prediction head to output $N$ adjacent vertices predictions $P = \{p_i \mid i = 1, 2, ..., N\}$, 
	$p_i \in \mathbb{R}^{4}$, 
	which contain relative coordinates and class probability of road and non-road.
    Given that an original input image can be relatively large, a sliding-window strategy is employed to divide the image into smaller patches.
    Under this setting, each candidate vertex in the set $C$
    receives $N$ predictions from every patch that covers it.
    Specifically, for a candidate vertex $v$, it accumulates $N \times K$ predictions, where $K$ denotes the number of overlapping patches containing $v$. Since the predictions within each patch are generated by the same set of $N$ learned queries with a consistent ordering, the outputs from different patches can be directly averaged into 
    $N$ aggregated predictions, eliminating the need for explicit point-to-point matching across patches.
	Finally, the valid adjacent vertices are selected out by a threshold $T_{valid}$.

  \textbf{Flexibility.}
    While our adjacency prediction strategy shares conceptual similarities with Sat2Graph \cite{he2020sat2graph}, it introduces key differences.
    Our decoupled strategy assigns dedicated modules for vertex detection and adjacent-vertex prediction, whereas the tightly coupled strategy of Sat2Graph forces two tasks of different nature (classification and regression) to share the same network architecture. Therefore, our AVP enables independent adjacency prediction for each candidate vertex, offering greater flexibility. In contrast, Sat2Graph performs vertex detection and adjacency prediction across all pixels within an image patch.

  \textbf{Efficiency.}
    Our cropping strategy fundamentally diverges from RNGDet \cite{xu2022rngdet, xu2023rngdet++}. 
    While RNGDet performs sequential image cropping with full model forward inference at each adjacent vertex prediction, 
    we perform feature interpolation on the visual encoder's output features generated by a single backbone forward pass,
    and such local features are enough to predict high-quality adjacent vertices, thus significantly lowering the computation cost as demonstrated in Tab. \ref{tab:ablation_ROI}.

\begin{figure}[t]
  \centering
  \includegraphics[width=0.8\linewidth]{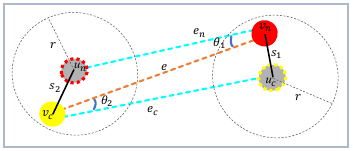}
   \caption{
   A schematic connecting current candidate vertex $v_n$ to another candidate $v_c$ (both from the detected candidate vertex set $C$), based on their respective adjacency predictions $u_n$ and $u_c$. The radius threshold $r$ defines the search range for locating an existing candidate vertex (e.g., $v_c$) that may correspond to a predicted adjacent vertex (e.g., $u_c$ corresponding to $v_n$, or $u_n$ corresponding to $v_c$). The predicted edges are denoted as $e_1 = {\{v_n, u_n\}}$ and $e_2 = {\{v_c, u_c\}}$, while $e = {\{v_n, v_c\}}$ represents the candidate edge to be verified. $s_1$ and $s_2$ denote the Euclidean distances between the possible vertex pairs ($v_n, u_c$) and ($v_c, u_n$), respectively. $\theta$ is the angle discrepancy between the predicted edges ($e_1$/$e_2$) and the candidate edge $e$. 
   }
   \label{fig:equation_decoding}
\end{figure}

\subsection{Initial Graph Constructor} 
\label{sec:constructor}
IGC takes the candidate vertices $C$ and the adjacent vertex predictions $P$ for every $v_c$ in $C$
as inputs an outputs a road network graph $G$.
Since AVP predicts adjacent vertices independently for each candidate vertex, without directly modeling the connectivity among candidate vertices, the predicted adjacency does not perfectly align with the set of candidates obtained from CVD,
resulting in positional deviations (as shown in Fig. \ref{fig:decoding}a). 
We employ the handcrafted decoding algorithm from \cite{he2020sat2graph} to decode these predictions into edges connecting the candidate vertices, 
thereby constructing an initial coherent road network graph. 

Notably, all candidate vertex coordinates are defined in the global coordinate system of the original image and since the adjacent vertex predictions are represented in relative coordinates and have already been averaged across different patches, the graph structure can be easily constructed at the global level.

The decoding process is illustrated in Fig. \ref{fig:equation_decoding}.
Essentially, the algorithm performs bidirectional, prediction-based matching: two detected candidate vertices, $v_n$ and $v_c$, are connected if their predicted adjacent vertices ($u_n$ and $u_c$) mutually correspond within a predefined tolerance.
This correspondence is determined by minimizing the total discrepancy $z$, as defined in Eq. \ref{eq:decoding}. Specifically, $s_1$ and $s_2$ denote the Euclidean distances between each predicted adjacent vertex and the corresponding existing candidate vertex found within the search radius $r$ (i.e., $s_1 = \lVert v_n - u_c \rVert$, $s_2 = \lVert v_c - u_n \rVert$), while $\theta_1$ and $\theta_2$ represent the angular deviations between the predicted edges ($e_1 = \{v_n, u_n\}$, $e_2 = \{v_c, u_c\}$) and the candidate edge $e = \{v_n, v_c\}$.
The weighting factor $w$, which converts angular deviation into an equivalent Euclidean distance, is set to 10, following Sat2Graph \cite{he2020sat2graph}.
During training, we adopt the bipartite matching strategy, specifically the Hungarian algorithm \cite{kuhn1955hungarian} to 
match the predicted adjacent vertices to the ground-truth adjacent vertices, 
thus providing supervision.

\begin{equation}
    \begin{aligned}
    z = s_1 + s_2 +w\cdot (1-\cos(\theta_1)) + w\cdot (1-\cos(\theta_2))
    \end{aligned}
    \label{eq:decoding}
\end{equation}

\subsection{Graph Expander}
\label{sec:expander}
The initial graph construction in Sec. \ref{sec:constructor} relies solely on the detected candidate vertices, 
which makes the candidate vertex set fixed and may lead to the omission of some road vertices.
In our approach, we incorporate a graph expansion step to dynamically extend the initial graph and 
enhance the completeness of the road network. 
In general, vertices with a degree of 1 in the current graph are selected as candidates for expansion. In the expansion-only strategy (Tab. \ref{tab:ablation_strategy}), however, the first expansion step starts from vertices with a degree of 0, after which the process proceeds as usual.
These candidate vertices undergo the adjacent vertex prediction step outlined in Sec. \ref{sec:predictor}, 
but with a higher validity threshold applied (only when inserting new vertices). 
Valid predicted vertices previously absent from the current graph are directly inserted (see Fig. \ref{fig:expansion}a) while those close to existing ones are merged if within a distance threshold $D_{merge}$ (see Fig. \ref{fig:expansion}b) following graph-growing methods \cite{bastani2018roadtracer, tan2020vecroad, xu2022rngdet, xu2023rngdet++}.
This process, analogous to graph-growing methods, can be repeated until no further expansion is possible; however, in practice, we find that as few as 3 expansion iterations are sufficient to achieve satisfactory results, as mentioned in Sec. \ref{sec:ablation}.

Interestingly, starting from a substantial set of initial candidate vertices without any existing edge (e.g., in the expansion-only strategy in Tab. \ref{tab:ablation_strategy}, all candidate vertices in $C$ directly constitute the initial graph, where every vertex has a degree of 0),
the road network graph can be efficiently constructed through 
simple merging (adding edges) and vertex insertion (adding vertices and edges). 
This aligns closely with graph-growing algorithms but allows for simultaneous growth from multiple seeds.

Therefore, our method supports graph building in three different modes: graph-generating, graph-growing, and a hybrid combination of the two, as described in Alg. \ref{alg:deh4r}.

\begin{algorithm}[t]
\caption{Local Minima Localization}
\label{alg:localmin_math}
\KwIn{Mask $M \in \mathbb{R}^{h\times w}$, threshold $\tau$}
\KwOut{Indices of detected vertices $\mathcal{I}$}

$A \gets -M$

$N \gets define\_8connected\_neighborhood()$

\tcp{$N \gets \{(i,j) \mid |i| \le 1, \, |j| \le 1\}$}

\ForEach{$(i,j)$}{
    \eIf {$is\_minimum(A_{i,j}, N)$}{
        $L_{i,j} \gets 1$
    }{
        $L_{i,j} \gets 0$
    }
}

$B \gets (A = 0)$ \tcp{zero-valued regions}

$B' \gets binary\_erosion(B, N)$ 

\tcp{pixels outside boundary assumed $1$}

$D \gets L \wedge B'$ \tcp{candidate vertex map}

$\mathcal{I} \gets extract\_indices(D, M, \tau)$

\tcp{$\mathcal{I} \gets \{(i,j) \mid D_{i,j}=1 \text{ and } M_{i,j} > \tau\}$}

\Return{$\mathcal{I}$}
\end{algorithm}

\begin{algorithm}[t]
\caption{DeH4R}
\label{alg:deh4r}
\KwIn{An aerial image $I$, graph building mode $mode$, expansion limit $exp\_limit$}
\KwOut{A road network graph $G = (V, E)$}

$tiles \gets sliding\_window\_partition(I)$

$M, F \gets execute\_CVD\_tile\_by\_tile(tiles)$

$C \gets find\_candidate\_vertices(M)$ 

\If{$mode =$ decoding \textbf{or} $mode =$ combination}{
    $P \gets infer\_adj\_tile\_by\_tile(C, F, tiles)$ 
    
    $G \gets decoding(C, P)$
    
    \If{$mode =$ combination}{
        $num\_exp \gets 0$ 
        
        \While{$num\_exp < exp\_limit$}{
            $S \gets find\_degree\_one\_vertices(G)$ 
            
            \If{$|S| = 0$}{
                break
            }
            
            $P \gets infer\_adj\_tile\_by\_tile(S, F, tiles)$ 
            
            $G \gets update\_graph(G, P)$ 
            
            $num\_exp \gets num\_exp + 1$ 
        }
    }
}
\ElseIf{$mode =$ expansion}{
    $num\_exp \gets 0$ 

    $S \gets C$ 

    \While{$num\_exp < exp\_limit$}{

        \If{$|S| = 0$}{
            break
        }

        $P \gets infer\_adj\_tile\_by\_tile(S, F, tiles)$

        $G \gets update\_graph(G, P)$

        $e  xp\_time \gets num\_exp + 1$

        $S \gets find\_degree\_one\_vertices(G)$
    }
}
\Return{$G$}
\end{algorithm}

\subsection{Loss}
	For a given image patch, DeH4R first predicts the keypoint heatmap $\hat{S_k}$, 
	the sampling point heatmap $\hat{S_s}$, and the road surface heatmap $\hat{S_r}$ first. 
    Based on the candidate vertices extracted from these heatmaps, the model then predicts potential adjacent vertices and corresponding class probabilities individually for all candidate vertices within the patch.
    Consequently, the overall loss is composed of three parts: segmentation loss $\mathcal{L}_{seg}$, class loss $\mathcal{L}_{class}$ and coordinate loss $\mathcal{L}_{coord}$.
	
	The segmentation loss is defined as:
	\begin{equation}
	\label{eq:key1}
	\begin{aligned}
	\mathcal{L}_{seg} = & \ \mathrm{BCE}(\hat{S}_k, S_k^*) + \mathrm{BCE}(\hat{S}_s, S_s^*) \\
	& + \mathrm{BCE}(\hat{S}_r, S_r^*)
	\end{aligned}
	\end{equation}
	where ${S_k}^*$ , ${S_s}^*$, and ${S_r}^*$ are the corresponding ground-truth masks.

    For each candidate vertex, the model predicts a set of potential adjacent vertices
    $\{p_i \mid i = 1, 2, ..., N\}, p_i \in \mathbb{R}^{4}$, each including probabilities of two classes (i.e. road \& non-road) and a pair of 2D coordinates.
    Here, we redefine them as $\{\hat{v}_i = (\hat{c}_i, \hat{n}_i) \mid i = 1, 2, ..., N\}$, with $\hat{c}_i \in \mathbb{R}^2$ representing the predicted class probabilities for two classes, and $\hat{n}_i \in \mathbb{R}^2$ representing the predicted 2D coordinates. The Hungarian algorithm optimally pair the predicted results with 
	the ground-truth set $\{v_j^* = (\hat{c}_j, \hat{n}_j)\mid j = 1, 2, ..., M\}$. 
	For a specific pair of vertices, the class loss and coordinate loss are defined as:

	\begin{equation}
		\label{eq:key2}
		\mathcal{L}_{coord}(\hat{v}_i,v_{\sigma(i)}^*) = \parallel\hat{n}_i-n_{\sigma(i)}^*\parallel
	\end{equation}

  	\begin{equation}
		\label{eq:key3}
		\mathcal{L}_{class}(\hat{v}_i,v_{\sigma(i)}^*) = \mathrm{CE}(\hat{c}_i,c_{\sigma(i)}^*)
	\end{equation}

  where $\parallel \cdot \parallel$ denotes the L2 norm, and CE denotes Cross-Entropy Loss.
  $\sigma$ indicates the optimal matching determined by the Hungarian algorithm.
 
  Notably, the class loss is computed for all predicted vertices and averaged over the entire prediction set, assigning a non-road label to unmatched vertices,
  whereas the coordinate loss is calculated exclusively for successfully matched vertex pairs and averaged over the associated subset.
	
Ultimately, the total loss is given by:
 	\begin{equation}
		\label{key4}
		\mathcal{L}=\lambda_1\mathcal{L}_{seg}+\lambda_2\mathcal{L}_{class}+\lambda_3\mathcal{L}_{coord}
	\end{equation}
	with weights $\lambda_1=\lambda_2=1$, and $\lambda_3=10$ to bring their magnitudes to a commensurable scale.

\section{Experiments}
\label{sec:exp}

\subsection{Datasets}
\label{sec:datasets}
We conducted experiments on two widely recognized public benchmarks, CityScale \cite{he2020sat2graph} and SpaceNet \cite{van2018SpaceNet}, to evaluate the performance of our method.
The CityScale dataset consists of 180 aerial images, each with a spatial resolution of 1 m and an image size of $2048 \times 2048$ pixels. Following the standard protocol, the dataset is divided into 144 images for training, 9 for validation, and 27 for testing. The SpaceNet dataset was originally released with images of $1300 \times 1300$ pixels at a ground resolution of approximately 0.3 m. To ensure consistency and enable fair comparisons across methods, all images are resampled to a uniform ground resolution of 1 m, resulting in a standardized image size of $400 \times 400$ pixels. SpaceNet contains 2,549 images in total, split into 2,040 for training, 127 for validation, and 382 for testing.

Beyond their technical specifications, both datasets provide not only high-resolution aerial imagery but also vector annotations of road centerlines, which serve as reliable ground-truth references for supervised learning. Moreover, they cover a wide geographic range and include a broad variety of scene types, from dense complex road networks to more sparse and simpler topologies. These diverse characteristics make the CityScale and SpaceNet datasets highly representative and widely adopted in the literature, and they have therefore become standard benchmarks for both training and evaluating algorithms designed for road network extraction.

\subsection{Metrics}
\label{sec:metrics}
We adopt TOPO \cite{biagioni2012inferring} (including precision, recall, and the composite metric F1), 
Average Path Length Similarity (APLS) \cite{van2018SpaceNet}, 
and Intersection over Union (IoU) for quantitative evaluation. 
TOPO primarily focuses on the existence and connectivity of edges, 
while APLS considers both topological connectivity and the spatial accuracy of the road network graph. 
IoU, on the other hand, 
measures the spatial alignment between the predicted road network graph and the ground-truth annotations by taking a buffer around the road network graph.

Formally, let the ground-truth graph be denoted as $G^*$ and the predicted graph as $\hat{G}$, the APLS is defined as follows:

\begin{equation}
    \begin{aligned}
    APLS = \frac{1}{M} \sum_{i=1}^{M} \frac{2}{\frac{1}{D_{\hat{G} \to G^*}} + \frac{1}{D_{G^* \to \hat{G}}}}
    \end{aligned}
    \label{eq:apls}
\end{equation}
where 
\begin{equation}
    \begin{aligned}
    D_{G^* \to \hat{G}} = 1-\frac{1}{N}\sum_{j=1}^{N} \min\left(1,\frac{|L(a,b)-L(\hat{a},\hat{b})|}{L(a,b)}\right)
    \end{aligned}
    \label{eq:apls_2}
\end{equation}
In Eq. \ref{eq:apls_2}, $N$ denotes the total number of unique paths considered, and $L(a, b)$ represents the length of the path connecting vertices $a$ and $b$. The summation is performed over all possible source–target vertex pairs $(a, b)$ within the ground-truth graph. For each ground-truth vertex $a$, the symbol $\hat{a}$ is used to denote its nearest counterpart in the predicted graph. In situations where a valid path between $\hat{a}$ and $\hat{b}$ does not exist, the evaluation assigns the maximum penalty value of 1.0, thereby ensuring that missing or disconnected routes are strongly penalized in the metric.
The definition of $D_{\hat{G}\to G^{}}$ follows the same formulation as in Eq. \ref{eq:apls_2}, with the roles of $G^{}$ and $\hat{G}$ interchanged.
In Eq. \ref{eq:apls}, $M$ denotes the total number of image samples in the dataset split.

In the TOPO metric, we start with a seed vertex pair $(a, a')$, where $a$ belongs to the ground-truth graph $G^*$ and $a'$ to the predicted graph $\hat{G}$, aligned in spatial location. Given a radius, let $P$ denote the set of vertices reachable from $a'$ within $\hat{G}$, and $Q$ the set of vertices reachable from $a$ within $G^{}$. We then define $C$ as the subset of vertices in $P$ and $Q$ that can be mutually matched under spatial correspondence. Then we define:

\begin{equation}
    Precision = \frac{|C|}{|P|}
    \label{eq:precision}
\end{equation}

\begin{equation}
    Recall = \frac{|C|}{|Q|}
    \label{eq:recall}
\end{equation}

\begin{equation}
F_1 = \frac{2 \cdot Precision \cdot Recall}{Precision + Recall}
\label{eq:f1}
\end{equation}
The final TOPO metrics are obtained by averaging over multiple sampled seed pairs.

\begin{table*}[!ht]
\caption{
    Quantitative evaluation results on the CityScale dataset and the SpaceNet dataset. 
    Best results are in bold and the second best are underlined.
    $^{\dag}$ means the result is reproduced with public codes.
    "-" means the corresponding results are not available.
    "*" denotes using ViT-B backbone from SAM \cite{kirillov2023segment}.
    The results of Seg-UNet, Seg-DRM, Seg-Improved, Seg-DLA, and  RoadTracer are taken from \cite{xu2023rngdet++}, 
    and the result of TD-Road is taken from \cite{he2022td}.
    }
    \centering
    \resizebox{\textwidth}{!}{
    \begin{tabular}{@{}lcccccccccc@{}}
        \toprule
        \multirow{2}{*}[-0.5ex]{Method} & \multicolumn{5}{c}{CityScale} & \multicolumn{5}{c}{SpaceNet}  \\
        \cmidrule(lr){2-6} 	\cmidrule(lr){7-11} 
        & Prec.$\uparrow$ & Rec.$\uparrow$ & F1$\uparrow$ & {APLS $\uparrow$} &  {IoU $\uparrow$}  & Prec. $\uparrow$ & Rec. $\uparrow$ & F1 $\uparrow$ &{APLS $\uparrow$}&{IoU $\uparrow$} \\  
        \midrule
        Seg-UNet  & 75.34 & 65.99 & 70.36 & 52.50 &--& 68.96& 66.32 & 67.61 & 53.77 &-- \\
        Seg-DRM  & 76.54 & 71.25 & 73.80 & 54.32&--& 82.79 & 72.56 & 77.34 & 62.26 &-- \\
        Seg-Improved & 75.83 & 68.90 & 72.20 & 55.34&--&81.56 & 71.38 & 76.13 & 58.82 &--\\
        Seg-DLA & 75.59 & 72.26 & 73.89 & 57.22 &--& 78.99& 69.80& 74.11 &56.36 &--\\							    
        RoadTracer  &78.00& 57.44&66.16&57.29&--	 &78.61&62.45&69.90&56.03&--\\
        Sat2Graph$^{\dag}$  & 80.15$^{\dag}$ &71.75$^{\dag}$&75.37$^{\dag}$&63.53$^{\dag}$&45.50$^{\dag}$		 	 &85.93&76.55&80.97&64.43&--\\
        TD-Road  & 81.94 & 71.63 &76.43&65.74&--    			 	 &84.81&77.80&81.15&65.15&--\\
        RNGDet++$^{\dag}$  	& 85.65$^{\dag}$ &72.58$^{\dag}$&78.44$^{\dag}$&67.72$^{\dag}$&45.61$^{\dag}$			 &91.37$^{\dag}$&75.24$^{\dag}$&82.52$^{\dag}$&67.81$^{\dag}$&40.70$^{\dag}$\\
        SAM-Road$^{\dag}$  & \textbf{91.27$^{\dag}$} & 65.70$^{\dag}$ & 76.20$^{\dag}$ & 66.54$^{\dag}$ & 33.87$^{\dag}$ 			&\textbf{91.48$^{\dag}$}&75.49$^{\dag}$&82.72$^{\dag}$&71.10$^{\dag}$&\underline{48.08$^{\dag}$}\\
        \midrule
        DeH4R*   & 83.25 & \underline{76.79} & \underline{79.71} & \underline{70.97} & \underline{54.94}			&89.01&\underline{81.69}&\underline{85.20}&\underline{72.68}&47.33\\
        DeH4R  & 82.50 & \textbf{80.18} & \textbf{81.21} & \textbf{72.38} & \textbf{55.79}			&86.22&\textbf{85.86}&\textbf{86.04}&\textbf{73.37}&\textbf{48.54}\\ 
        \bottomrule
        \end{tabular}
    }
    \label{tab:quantitative}

\end{table*}

\begin{table}[!t]
\setlength{\tabcolsep}{2pt}
    \caption{Quantitative results of different backbones on the CityScale dataset. Vit-B refers to the pretrained backbone in SAM \cite{kirillov2023segment}.
    }
    \centering
    \resizebox{\columnwidth}{!}{
      \begin{tabular}{lcccccc}
      \toprule
      Method & Backbone & Prec.$\uparrow$ & Rec.$\uparrow$ & F1$\uparrow$ & APLS $\uparrow$ & IoU $\uparrow$\\ 
      \midrule
      Sat2Graph & DLA-34 & 80.15 & 71.75 & 75.37 & 63.53 & 45.50 \\
      \rowcolor{gray!30}
      Sat2Graph & ViT-B & 83.08 & 69.36 & 75.64 & 67.55  & 52.72 \\

      RNGDet++  & R-101 & 85.65 & 72.58 & 78.44 & 67.72 & 45.61 \\
      \rowcolor{gray!30}
      RNGDet++  & ViT-B & 84.82 & 71.95 & 77.68 & 65.61 & 43.26 \\

      SAM-Road & R-101 & 90.13 & 67.14 & 76.73 & 65.89 & 48.22 \\
      \rowcolor{gray!30}
      SAM-Road & ViT-B & 91.27 & 65.70 & 76.20 & 66.54 & 33.87 \\

      DeH4R   & R-101 & 77.93 & 81.61 & 79.61 & 69.92	& 49.98 \\ 
      \rowcolor{gray!30}
      DeH4R   & ViT-B & 83.25 & 76.79 & 79.71 & 70.97	& 54.94 \\ 
      \bottomrule
    \end{tabular}
    }
    \label{tab:backbone}
\end{table}

\subsection{Implementation Details}
	\textbf{Model.} We adopt the SAM2 \cite{ravi2024sam} Hiera-B+   as the Visual encoder
	and the output dimension is 256.
	In AVP, the hidden dimension of Transformer Decoder layer is 128, 
	the number of attention heads is 8, and the dimension of FFN is 256.

	\textbf{Train.} AdamW optimizer is adopted with base learning rate (lr) of 0.001, 
	and lr is decayed by 0.1 at milestones of 7, 11, 15 on CityScale
	and 10, 20, 25 on SpaceNet.
  	For both datasets, $N_q$ is 10, $l$ is 3.
	We apply random 90-degree rotations to a training patch and add a Gaussian noise $\mathcal{X}\sim N(0,1)$ to adjacent vertices. 
	Every ground-truth vertex within a patch and its direct neighbors constitute a sample,
	and we shuffle the samples in the patch and select the first $N_{points}$ of them for AVP training, 
	with empty samples masked out. 
	On CityScale, $N_{points}$ is 220, patch size is 512$\times$512, epoch is 20.
	On SpaceNet, $N_{points}$ is 64, patch size is 256$\times$256, epoch is 30.
	Following \cite{hetang2024segment}, 
	we fine-tuned the pre-trained Hiera-B+ with $\times$ 0.1 base lr to adapt it to our task. 
    
    Notably, in order to guarantee a fair comparison, our reproduction of the SAM-Road \cite{hetang2024segment} method was trained strictly on the designated training set. This differs from the original implementation, where both the training and validation sets were used jointly for training.

	\textbf{Inference.} For both datasets, $D_{merge}$ is empirically set to 10 pixels and the expansion limit is set to 3 times.
	On CityScale, inference batch size is 4, $T_{valid}$ is 0.5, $T_{valid}$ is 0.7 in expansion steps.
	On SpaceNet, inference batch size is 16, $T_{valid}$ is 0.45, $T_{valid}$ is 0.65 in expansion steps.
    The edges of predicted graph and the ground-truth graph are 
rasterized as 3-pixel wide lines to compute the IoU score.

	All experiments were conducted on a server equipped with 8$\times$3090 GPUs. For each specific training run, only 4 GPUs were utilized, and all inference results produced by us were obtained on the same GPU (GPU \#0).

\subsection{Main Results}
  \label{sec:results}
    We selected four segmentation-based methods (Seg-UNET \cite{ronneberger2015u}, Seg-DRM \cite{mattyus2017deeproadmapper}, Seg-Improved \cite{batra2019improved}, and Seg-DLA \cite{yu2018deep})
    two graph-growing algorithms (RoadTracer \cite{bastani2018roadtracer} and RNGDet++ \cite{xu2023rngdet++}), 
    and three graph-generating methods (Sat2Graph \cite{he2020sat2graph}, TD-Road \cite{he2022td}, and SAM-Road \cite{hetang2024segment})  as baseline methods.
    The quantitative results on the CityScale and SpaceNet datasets are presented in Tab. \ref{tab:quantitative}. 
    Meanwhile, the results of experiments comparing graph-based methods with different backbones are presented in Tab. \ref{tab:backbone}.
    Additionally, the inference efficiency of the two best-performing baselines was also assessed, 
    with details provided in Tab. \ref{tab:speed_comparison}.
    
    DeH4R achieves new SOTA performance on both datasets, 
    leading in TOPO-F1, APLS, and IoU compared to other methods. 
    Specifically, DeH4R outperforms the second-best RNGDet++ on the CityScale dataset by \textbf{4.66} APLS, \textbf{2.77} TOPO-F1, 
    and \textbf{10.18} IoU, 
    and outperforms the second-best SAM-Road on the SpaceNet dataset by \textbf{2.27} APLS, \textbf{3.32} TOPO-F1, and \textbf{0.46} IoU.  Segmentation-based methods perform the worst, with APLS metrics more than \textbf{10} points lower than our method. 
    DeH4R achieves a relatively lower Topo-precision compared to some baseline methods. However, this should not be interpreted as a deficiency of our approach; rather, it reflects a normal precision–recall trade-off that arises naturally in graph-based road network extraction. This observation indicates that our decoupled design encourages greater exploration by the network and leads to a more balanced trade-off between precision and recall, achieving an improved overall performance.

  Compared to the current SOTA method, 
  DeH4R is approximately 10 times faster than graph-growing method RNGDet++ and 
  is on par with the most efficient graph-generating method SAM-Road in inference speed (see Tab. \ref{tab:speed_comparison}). 

  Notably, DeH4R demonstrates a more pronounced performance advantage on the CityScale dataset compared to SpaceNet.
  We attribute this to a “scale effect” introduced by the substantial difference in image sizes ($2048^{2}$ vs. $400^{2}$), which results in notable differences in the overall scale of the road networks covered by each image under the same spatial resolution. Since the APLS metric is normalized by the shortest path length between sampled node pairs, deviations of the same magnitude (in direction or distance) will incur relatively heavier penalties on the SpaceNet dataset, where the overall graph extent is smaller. In contrast, the TOPO metric is computed based on local neighborhoods and is less sensitive to the global graph scale; therefore, the TOPO improvement margins are comparable across the two datasets. For segmentation metrics (i.e., IoU), a similar scale-related effect also influences the observed gains.
  For evaluating the quality of long and consecutive road network graphs, 
  conclusions drawn from the CityScale dataset are empirically more reliable.
  Therefore, we conduct extra comparative experiments (no hyperparameter tuning for all variants) of graph-based methods with different backbones on CityScale for further comparison and the results are provided in Tab. \ref{tab:backbone}. 
  The results indicate that our method outperforms even when using weaker SAM-ViT-B and R-101 backbones. Compared with the tightly coupled method Sat2Graph, our approach substantially improves recall while maintaining comparable precision, demonstrating that the decoupled strategy effectively enhances the network’s exploratory capability and validating the effectiveness of its design.
	
\begin{figure*}[t]
		\includegraphics[width=\linewidth]{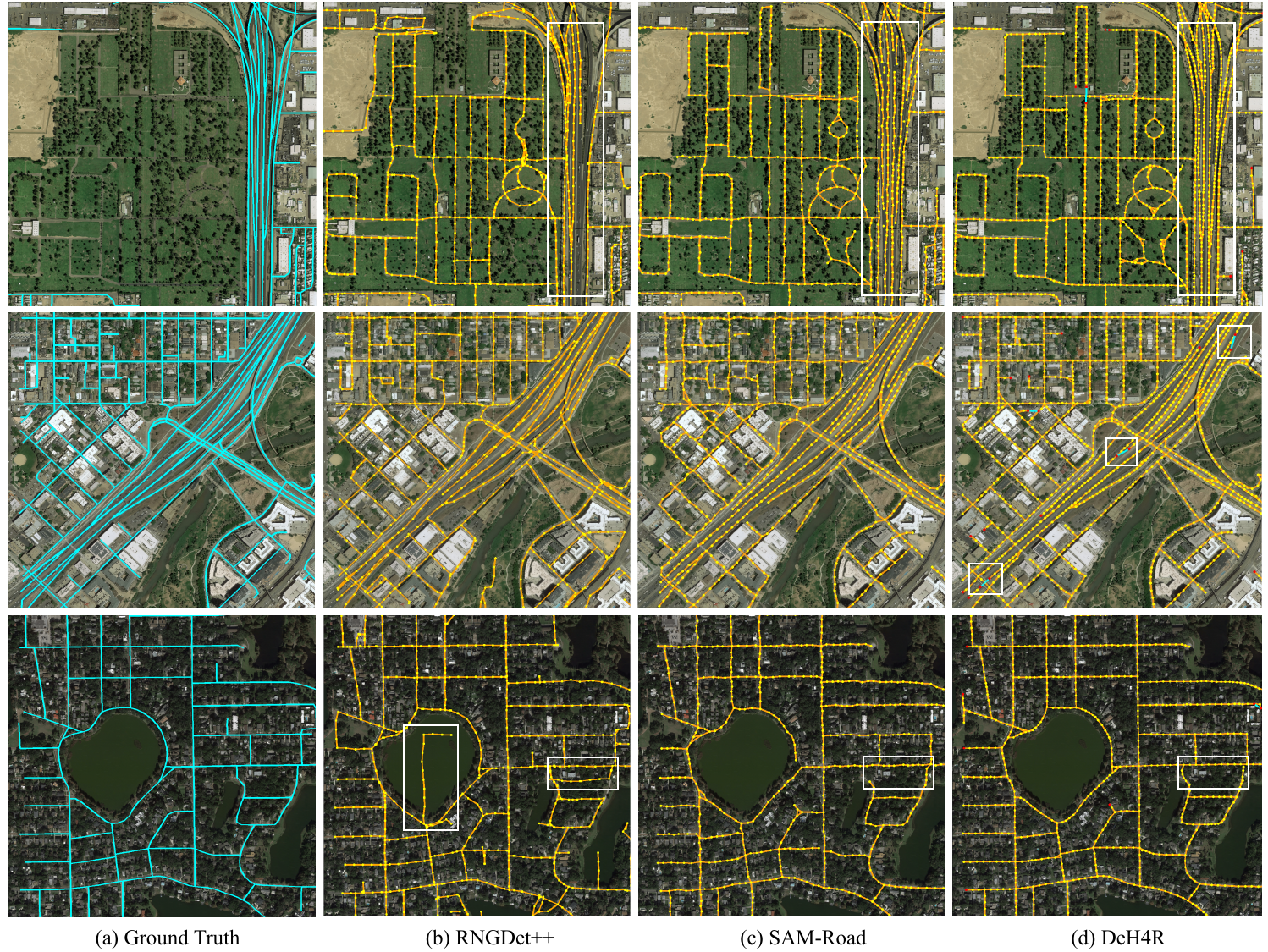}
		\caption{Qualitative visualizations.
		(a) Ground-truth road network graphs (cyan lines). 
		(b)-(d) Results of RNGDet++, SAM-Road, and our DeH4R.
		Yellow points represent vertices and 
		orange lines represent edges,
		with red points and blue lines in (d) denoting inserted vertices and edges by expansion, respectively. 
		}
		\label{fig:comp}
	\end{figure*}

    Fig. \ref{fig:comp} presents some qualitative results of RNGDet++, SAM-Road, and our method. 
    SAM-Road shows significant vertex deviations from road centerlines, resulting in noisy outputs. 
    Both DeH4R and RNGDet++ achieve smoother results. 
    For long straight roads (the 1st and 2nd row), 
    RNGDet++ exhibits frequent missed detections while SAM-Road exhibits fragments;
    under occlusion (the 3rd row), RNGDet++ over-explores while SAM-Road fractures.
    Our method consistently maintains topological accuracy and road centerline alignment. 
    Notably, despite missing labels in the park area (the 1st row), 
    all methods generalize well to detect the road network graph.

    \begin{table}[t]
    \caption{The inference time on both CityScale and SpaceNet datasets. 
    DeH4R is  $\sim$10$\times$ faster than graph-growing method RNGDet++ while on par with graph-generating method SAM-Road.}
    \centering
    \resizebox{\columnwidth}{!}{
    \begin{tabular}{@{}lcc@{}}
        \toprule
        Method & CityScale Dataset & SpaceNet Dataset \\
        \midrule
        RNGDet++ & 136.0 min & 145.0 min \\
        SAM-Road & \textbf{10.34 min} & 18.53 min \\
        DeH4R & 13.34 min & \textbf{15.24 min} \\
        \bottomrule
    \end{tabular}
    }
	\label{tab:speed_comparison}
	\end{table}

  \begin{table}[t]
  \caption{Ablation study on different strategies to generate final road network graphs.}
		\centering
		\resizebox{\columnwidth}{!}{
		\begin{tabular}{@{}cc|cccc@{}}
		\toprule
		Decoding & Expansion & Topo-F1$\uparrow$ & APLS$\uparrow$ & IoU$\uparrow$ & Infer. Time\\
		\midrule
		\checkmark & \checkmark & 81.21 & 72.38 & 55.79  & 13.34\\
		\checkmark & 			& 81.02 & 71.72 & 56.42  & 12.05\\
				   & \checkmark & 79.18 & 68.70 & 55.27  & 18.33\\
		\bottomrule
		\end{tabular}
		}
		\label{tab:ablation_strategy}
		\end{table}

	\begin{table}[t]
        \caption{Ablation study on different expansion times.}
		\centering
		\resizebox{\columnwidth}{!}{
		\begin{tabular}{@{}c|cccc@{}}
		\toprule
		Num. Exp. & Topo-F1$\uparrow$ & APLS$\uparrow$ & IoU$\uparrow$ & Infer. Time \\
		\midrule
		1 & 81.14 & 71.62 & 56.01 & 12.55 min\\
		2 & 81.17 & 71.61 & 55.87 & 13.01 min\\
		3 & 81.21 & 72.38 & 55.79 & 13.34 min\\
		4 & 81.22 & 72.31 & 55.73 & 13.65 min\\
		5 & 81.24 & 72.41 & 55.70 & 13.99 min\\
		6 & 81.26 & 72.41 & 55.68 & 14.30 min\\
		7 & 81.26 & 72.28 & 55.66 & 14.82 min\\
		\bottomrule
		\end{tabular}
		}
		\label{tab:ablation_expansion}
	\end{table}

	\begin{table}[t]
        \caption{Ablation study on different ROI size.}
		\centering
		\resizebox{\columnwidth}{!}{
		\begin{tabular}{@{}c|cccccc@{}}
		\toprule
		ROI & Prec.$\uparrow$ & Rec.$\uparrow$ & F1$\uparrow$ & {APLS $\uparrow$} &  {IoU $\uparrow$} & Infer. Time\\
		\midrule
		1x1 & 79.28 & 83.19 & 81.07 & 71.54 & 54.98 & 13.87 min \\
		3x3 & 82.50 & 80.18	& 81.21 & 72.38 & 55.79 & 13.34 min \\
		5x5	& 79.15 & 83.07 & 80.96 & 71.43 & 54.93 & 14.83 min \\
		7x7	& 79.46 & 82.83 & 81.00 & 71.26 & 55.15 & 15.08 min \\
		9x9	& 78.95 & 83.23 & 80.91 & 71.71 & 54.97 & 15.05 min \\
		\bottomrule
		\end{tabular}
		}
		
		\label{tab:ablation_ROI}
		\end{table}

	\begin{table}[ht]
        \caption{Ablation study on backbones.
		"SAM" denotes using ViT-B backbone from SAM,
		"SAM2 w/ M." represents full Hiera-B+ backbone, 
		and "SAM2 w/o M." represents using only the 1/16 scale features of Hiera-B+.
		}
		\centering
		\resizebox{\columnwidth}{!}{
		\begin{tabular}{@{}c|ccccc@{}}
		\toprule
		Backbone & Prec.$\uparrow$ & Rec.$\uparrow$ & F1$\uparrow$ & {APLS $\uparrow$} &  {IoU $\uparrow$} \\
		\midrule
		SAM     & 83.25 & 76.79 & 79.71 & 70.97 & 54.94  \\
		SAM2 w/o M. 	 & 79.34 & 83.34 & 81.19 & 71.79 & 55.13  \\
		SAM2 w/  M.   & 82.50 & 80.18 & 81.21 & 72.38 & 55.79  \\
		
		\bottomrule
		\end{tabular}
		}
		\label{tab:ablation_backbone}
		\end{table}
  \subsection{Ablation}
  \label{sec:ablation}
We conduct ablation experiments to study the effects of the key design choices on the CityScale dataset.

\textbf{Decoding vs. Expansion vs. Decoding \& Expansion.} 
DeH4R is able to construct road network graphs via graph-generating (decoding), graph-growing (expansion), or both (expansion after decoding) as specified in Alg. \ref{alg:deh4r}.
Tab. \ref{tab:ablation_strategy} shows that decoding alone yields high-quality road network results, 
expansion-only outperforms RNGDet++ but still falls behind decoding, 
and the hybrid mode achieves the best topological performance, validating not only the effectiveness of our decoupled design but also the further performance gains brought by the integration of both strategies through vertex insertion.
Reasonably, more vertices introduced to complete the graph result in the slight drop of IoU score.

\textbf{Different Expansion Times.} 
For conventional graph-growing methods, the iteration usually terminates when no further vertices or edges can be inserted, or when a predefined maximum number of iterations (e.g.,, 3000) is reached.
In contrast, our approach supports parallel expansion from multiple candidate vertices, enabling DeH4R to grow the graph more efficiently and thus reducing the required number of iterations.
As shown in Tab. \ref{tab:ablation_expansion}, our method performs graph expansion very efficiently, incurring minimal computational overhead even when multiple candidate vertices are expanded in parallel. However, this does not imply that increasing the number of expansion iterations will necessarily lead to better performance. Empirically, we observe that performing 3 rounds of graph expansion is sufficient to achieve satisfactory results. Further increasing the number of iterations beyond this point yields negligible performance gains, may even slightly degrade the quality of the generated road network and incurs computational overhead.

\textbf{Different ROI Size.} 
The prediction of adjacent vertices is neighborhood-based,
and the computational cost grows quadratically with the size of the ROI. 
In traditional graph-growing methods, adjacent vertex prediction is performed on ROIs cropped from the original image, requiring relatively large (e.g., $256 \times 256$) to capture sufficient context. In contrast, DeH4R extracts ROIs at the feature level, allowing much smaller regions while still preserving the necessary information.
The results in \ref{tab:ablation_ROI} indicate that using a $3\times3$ ROI is sufficient and efficient
to achieve high-quality road network graphs with short runtime.
It can also be observed that the precision–recall trade-off is not obtained smoothly as the ROI size increases, and we speculate that the interpolation and cropping performed on highly abstract features can introduce disturbances and a $3\times3$ ROI represents a potential sweet spot: smaller ROIs lack sufficient information, whereas larger ROIs may introduce additional noise.
Interestingly, processing $1\times1$ ROI can be slower than $3\times3$ ROI, possibly due to reduced parallelism and less efficient memory access on the GPU.

\textbf{SAM/SAM2 Backbones \& Multi-scale Features.}
For the backbones we employ, namely SAM-ViT-B and SAM2-Hiera-B+, although both have a comparable number of parameters, the latter has been trained on a larger dataset and provides multi-scale feature representations. To investigate the performance differences attributable to the backbone and the inclusion of multi-scale features, we compare SAM-ViT-B, SAM2-Hiera-B+, and SAM2-Hiera B+ without the multi-scale strategy.
We discard the multi-scale features generated by SAM2-Hiera-B+ backbone 
and use only the 1/16 scale features as a variant. 
The results in Tab. \ref{tab:ablation_backbone} show that 
the SAM2-Hiera-B+ backbone without multi-scale features brings a boost of 1.48 TOPO-F1 and 0.82 APLS compared to the SAM-ViT-B backbone,
and multi-scale features lead to a further improvement of 0.59 APLS and 0.66 IoU.
This indicates high-resolution features are important for accurate localization of small road vertices.

\subsection{Discussions and Limitations}
	\label{sec:limitation}
\begin{figure}[t]
      \centering
      \includegraphics[width=\linewidth]{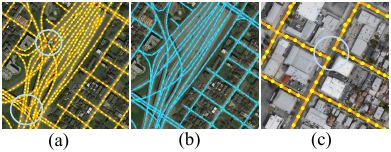}
       \caption{Examples of failure cases. (a) The overlapped road. (b) The label for (a). (c) Missed connection between two sharp-turn vertices which are supposed to be vertices with a degree of 3 (T-junction).}
       \label{fig:failure}
    \end{figure}

    Regarding the limitations of our method, false connections can occur in complex scenes, such as regions with overlapping roads (see Fig. \ref{fig:failure}a).
    This primarily results from the handcrafted decoding process employed in our approach.
    Meanwhile, our graph expansion procedure is specifically designed to recover missed vertices and edges (see Fig. \ref{fig:expansion}a and Fig. \ref{fig:expansion}b), and therefore cannot correct erroneous connections that already exist, and it may behave suboptimally in complex areas (see Fig. \ref{fig:failure}a) or sharp-turn vertices (see Fig. \ref{fig:failure}c).
    This limitation stems from the intrinsic design of our graph-growing strategy, which expands the graph only from existing vertices that have a degree of 1.
    An interesting and promising direction for future research would be to develop end-to-end frameworks that are capable of jointly learning both the initial graph construction and subsequent refinement. Such approaches may potentially overcome the current limitations by enabling the model to correct false connections dynamically while still recovering missed edges, thereby further improving the overall quality and topological accuracy of the generated road network graphs.

\section{Conclusion}
\label{sec:Conclusion}
In this study, we propose to decouple the complex problem of road network graph extraction into four distinct but interconnected stages, namely candidate vertex detection, adjacent vertex prediction, initial graph construction, and graph expansion, and we introduce DeH4R as an instantiation of this approach.
By explicitly separating these stages, DeH4R is able to leverage the respective advantages of both graph-generating and graph-growing methodologies. In particular, it maintains high computational efficiency while simultaneously allowing for dynamic vertex insertion, a feature that is often absent in previous methods. 
Experimental results on the widely used CityScale and SpaceNet benchmarks demonstrate that DeH4R achieves state-of-the-art performance across multiple evaluation metrics, validating both the efficacy of the decoupling strategy and the benefits of the proposed hybrid design.
We believe that DeH4R can serve as a robust and valuable benchmark for future research in the area of road network graph extraction, providing a foundation for the development of more advanced and accurate graph-based methods.


%

\ifCLASSOPTIONcaptionsoff
  \newpage
\fi



%
\bibliographystyle{IEEEtran}
\bibliography{main}

%





\begin{IEEEbiography}
[{\includegraphics[width=1in,height=1.25in,clip,keepaspectratio]{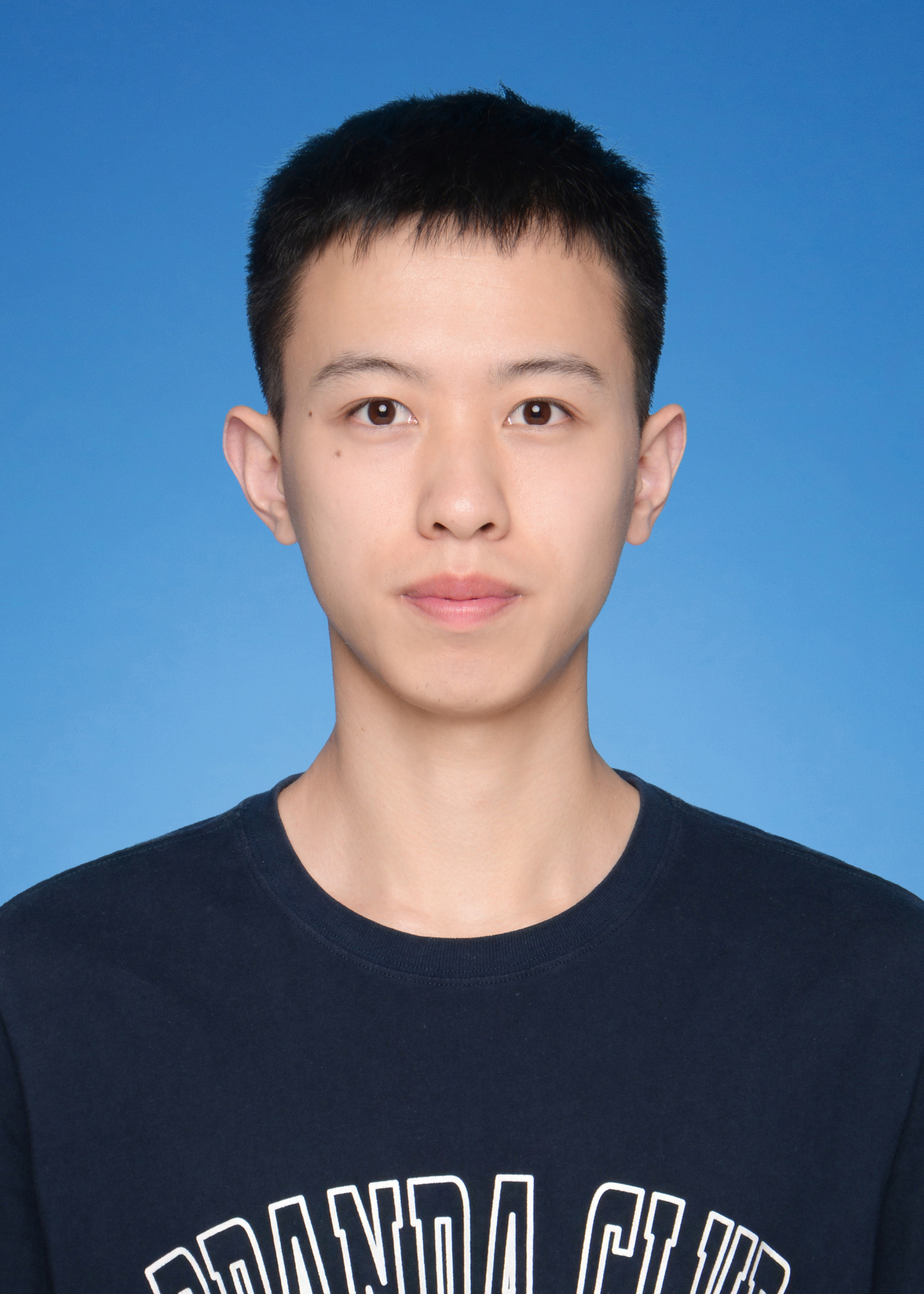}}]{Dengxian Gong}
received the B.S. degree from Chongqing University, Chongqing, China, in 2023. He is currently pursuing the Master's degree with Wuhan University, Wuhan, China. His research interests include image understanding, multimodal large language models, and remote sensing.
\end{IEEEbiography}

\begin{IEEEbiography}[{\includegraphics[width=1in,height=1.25in,clip,keepaspectratio]{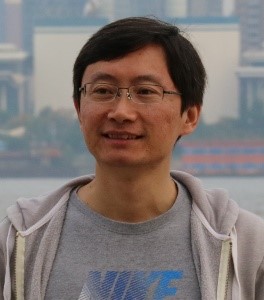}}]{Shunping Ji} (Senior Member, IEEE) received the Ph.D. degree in photogrammetry and remote sensing from Wuhan University, Wuhan, China, in 2007. He is currently a Professor with the School of Remote Sensing and Information Engineering, Wuhan University. His research interests include photogrammetry, remote sensing image processing,
mobile mapping system, and machine learning.

\end{IEEEbiography}




\end{document}